\documentclass{article}

\usepackage{microtype}
\usepackage{graphicx}
\usepackage{subfigure}
\usepackage{booktabs}
\usepackage{amsmath}
\usepackage{amsthm}
\newtheorem{theorem}{Theorem}
\usepackage{amsfonts}
\usepackage{enumitem}
\usepackage[numbers]{natbib}

\usepackage{hyperref}

\usepackage[accepted]{icml2019_phys4dl}

\icmltitlerunning{Scale Steerable Filters}

\begin{document}

\twocolumn[
\icmltitle{Scale Steerable Filters for Locally Scale-Invariant Convolutional Neural Networks}




\begin{icmlauthorlist}
\icmlauthor{Rohan Ghosh}{to}
\icmlauthor{Anupam K. Gupta}{to}

\end{icmlauthorlist}

\icmlaffiliation{to}{National University of Singapore, Singapore}
\icmlcorrespondingauthor{Rohan Ghosh}{rghosh92@gmail.com}

\icmlkeywords{Machine Learning, ICML}

\vskip 0.3in
]



\printAffiliationsAndNotice{}  

\begin{abstract}
 Augmenting transformation knowledge onto a convolutional neural network's weights has often yielded significant improvements in performance. For rotational transformation augmentation, an important element of recent approaches has been the use of filters exactly steerable in their rotation, using circular harmonics. Here, we propose a scale-steerable filter basis for the locally scale-invariant CNN, denoted as log-radial harmonics.  By replacing the kernels in the locally scale-invariant CNN \cite{lsi_cnn} with scale-steered kernels, significant improvements in performance can be observed on the MNIST-Scale and FMNIST-Scale datasets. Training with a scale-steerable basis results in a) filters which show meaningful structure, and b) feature maps which demonstrate visibly higher spatial-structure preservation of input. The proposed scale-steerable CNN shows on-par generalization with affine transformation estimation methods such as Spatial Transformers, in response to test-time data distortions.
\end{abstract}

\section{Introduction} \label{sec:intro}

Convolutional Neural Networks rise to success on large datasets like ImageNet in \cite{alexnet}, has prompted a myriad of work in their direction, which build on their key depth-preserved transformation equivariance property to achieve better classifiers \cite{g_cnn,rotation_steer_gcnn,rotation_vectorfields}. Equivariance to transformations has been thus recognized as an important pre-requisite to any classifier, and CNNs which are by definition translation equivariant have been recognized as a first important step in this direction. 

An underlying requirement to a transformation equivariant representation is the construction of transformed copies of filters, i.e. when the transformation is a translation, the operation becomes a convolution. A natural extension of this idea to general transformation groups led to the idea of Group-equivariant CNNs \cite{g_cnn}, where in the first layer, transformed copies of filter weights are generated. Subsequently, the application of group convolution ensures that the network stays equivariant to that transformation throughout. 

However, there are certain issues pertaining to the application of any (spatial) transformation on a filter:
\setlist{nolistsep}
\begin{enumerate}
    \item There is no prior on the spatial complexity of a convolutional filter within a CNN, which means a considerable part of the filter space may contain filters which are not sensitive to the desired spatial transformation. Examples include rotation symmetric filters, high-frequency filters etc.    
    \item As noted in \cite{rotation_steer_gcnn}, most transformations are continuous in nature, necessitating interpolation for obtaining filter values at new locations. This usually leads to interpolation artifacts, which can have a greater disruptive effect when the filters are usually of small size.  
\end{enumerate}
\paragraph{Steerable Filters}

To alleviate these issues, the use of a \textit{steerable} filter basis for filter construction and learning was proposed in \cite{steerable_cnn}. Steerable filters have the unique property, that allow them to be transformed by simply using linear combinations of an appropriate steerable filter basis. Importantly, the choice of the steerable basis allows one to control the transformation sensitivity of the final computed filter. Especially for a circular harmonic basis \cite{hnet}, we find that filters of order $k$ are only sensitive to rotation shifts in the range $(0,2\pi/k)$. In this case, higher order filter responses show less sensitivity to input rotations, and simultaneously are of higher spatial frequency and complexity. Using a small basis of the first few filter orders enabled the authors of \cite{rotation_steer_gcnn} to achieve state-of-the-art on MNIST-Rot classification (with small training data size). 

\section{Contributions of this Work} 

\paragraph{Log-Radial Harmonics: A scale steerable basis}
In this paper, we define filters which are steerable in their spatial scale using a complex filter basis we denote as log-radial harmonics. Each kernel of a CNN is represented as the real part of the linear combination of the proposed basis filters, which contains filters of various orders, analogous to circular harmonics. Furthermore, the scale steerable property permits exact representation of the filters in its scale simply through a linear combination of learnt complex coefficients on the log-radial harmonics. The filter form is conjugate to the circular harmonics, with the choice of filter order having a direct impact on the scale sensitivity of the resulting filters.
\paragraph{Scale-Steered CNN (SS-CNN)}
Using the log-radial harmonics as a complex steerable basis, we construct a locally scale invariant CNN, where the filters in each convolution layer are a linear combination of the basis filters. For obtaining filter response across scales, each filter is simultaneously steered in its scale and size, and the filter responses are eventually max-pooled. We demonstrate accuracy improvements with the scale-steered CNN on datasets containing global (MNIST-Scale, and FMNIST-Scale) and local (MNIST-Scale-Local; synthesized here) scale variations. Importantly, we find that on MNIST-Scale, the proposed SS-CNN achieves competitive accuracy to the Spatial Transformer Network \cite{st_network}, which due to its global affine re-sampling property has a natural advantage in this task. 

\section{Related Work} 

\paragraph{Previous work with Local Scale Invariant/Equivariant CNNs} 
Scale-transformed weights were proposed in \cite{lsi_cnn}, where it was observed to improve performance over the normal baseline CNN, on MNIST-Scale. On the same dataset (with a 10k, 2k and 50k split), better performance was observed in \cite{scale_vector_fields}, where in addition to forwarding the maximum filter response to a range of scales, the actual scale at which the response was obtained was also forwarded. In both works, weight scaling was only indirectly emulated, by rather scaling the input and the resizing back the convolution response to a fix size for max-pooling across scales.

\section{Background:Steerable Filters for Rotation}

Rotation steerable filters, in the form of circular harmonics, are of the form $W(r,\phi) =  R(r)F(\phi)$, expressed in polar co-ordinates. For circular harmonics, $R(r)$ is usually considered to be a Gaussian function centered on a particular radius. $F(\phi)$ is a complex function of unit norm, $e^{i(k\phi +\beta)}$. Such a choice of $F(\phi)$ allows one to rotationally steer the filter $W(r,\phi)$ by any angle $\theta$, just by a complex multiplication, $W(r,\phi+\theta)=W(r,\phi)e^{ik\theta}$. Furthermore, control over the rotational order $k$ allows one to directly control rotational sensitivity of the resulting filter (which is invariant to the filter rotation), and also simultaneously the spatial complexity of the filter.






\section{Methods} 
\subsection{Scale-steerable filters: Log-Radial Harmonics}
\begin{figure}\label{fig:steerable_basis}
  \centering
  \includegraphics[width=0.5\textwidth]{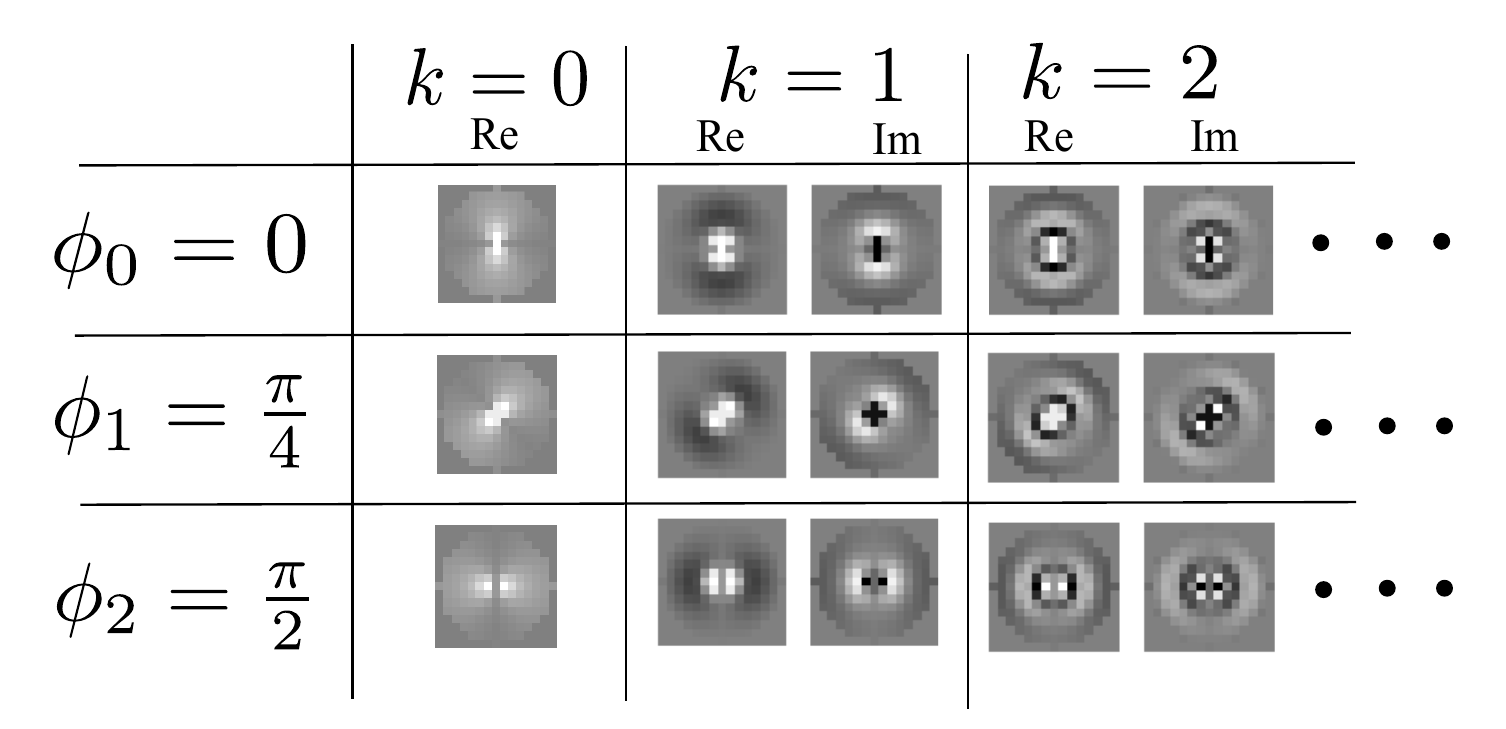}
  \caption{Scale-steerable basis filters for selected orientations and filter orders and $m=1$. Notice the centrality and log-polar nature of the basis filters.}
\end{figure}

 Similar to the rotation steerable circular harmonics, we can analogously construct a set of filters of the form $W(r,\phi)=\Phi(\phi)F(\log r)/r^m$. Since we wish to steer the scale of the filter, now $\Phi$ is of Gaussian form, whereas $F(\log r)$ is complex valued with unit norm, i.e. $e^{i(k\log r + \beta)}$. The proposed mathematical form of a scale steerable filter of order $k$ and centered on a particular $\phi=\phi_j$ is, 
\begin{equation}\label{eq:steerable_form} 
     S^{kj}(r,\phi) =\frac{1}{r^m} \left(K(\phi,\phi_j)+K(\phi,\phi_j+\pi)\right)e^{i(k(\log r)+\beta)}, 
\end{equation}

where $K(\phi,\phi_j) = e^{-d(\phi,\phi_j)^2/2\sigma_{\phi}^2}$. Here $d(\phi,\phi_j)$ is the distance between the two angles $\phi$ and $\phi_j$. Example filters constructed using equation \ref{eq:steerable_form} are shown in Figure \ref{fig:steerable_basis}. When steering the above filter in scale, we find that a complex multiplication of $s^{m-2}e^{-i(klogs)}$ suffices, where $s$ is the scale factor change. This we prove in the following theorem.
\begin{figure*}[!ht]
  \centering
  \includegraphics[width=0.8\textwidth]{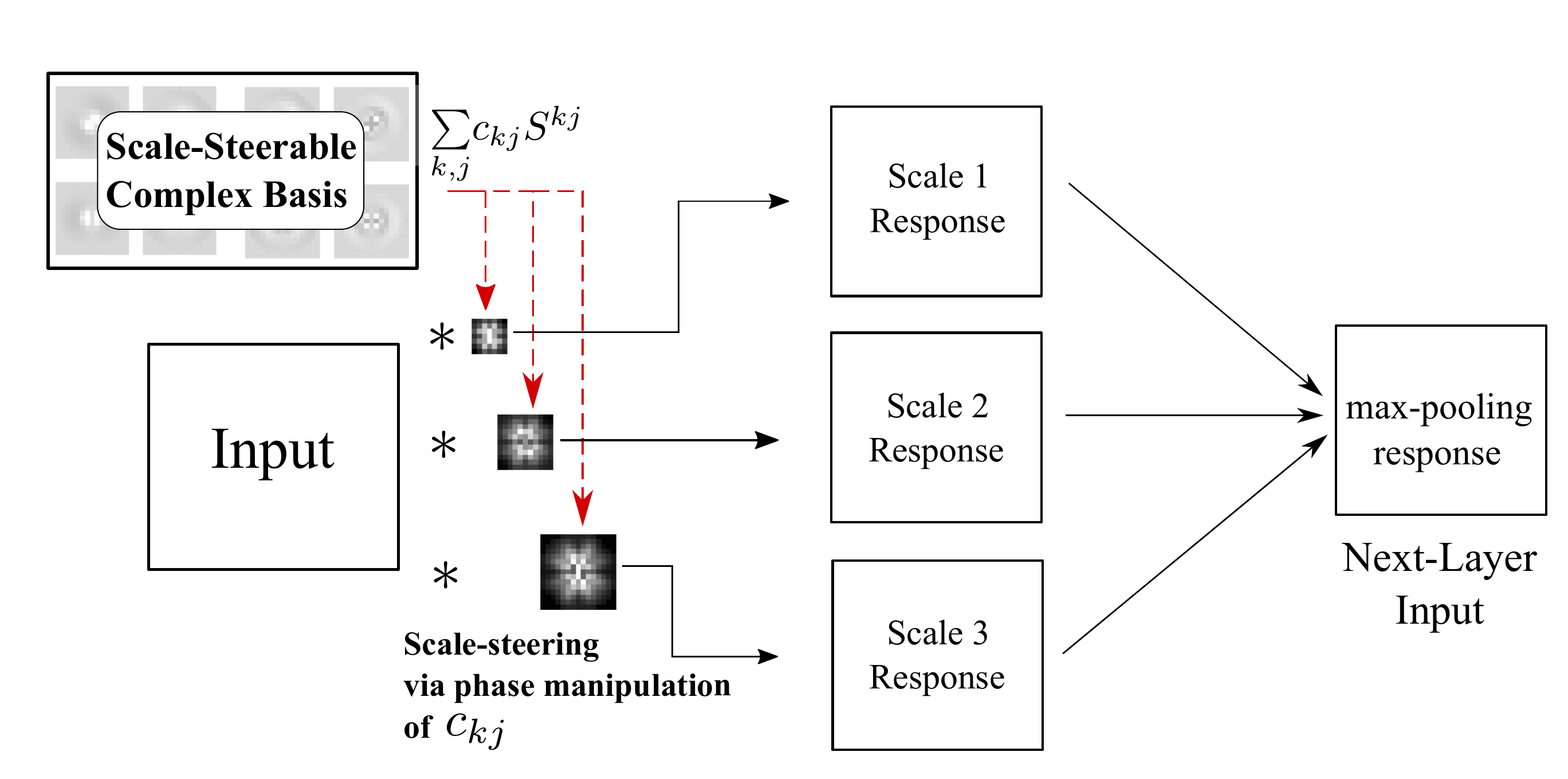}
  \caption{The proposed scale-invariant layer with scale-steered filters. The scaled versions of filter shown in this figure have been generated using phase manipulations over a scale-steerable basis (equation \ref{eq:scale_steering}). Note that the steerable basis is complex, but only the real part of the steered filter is used, ensuring that filter weights are real-valued.}
    \label{fig:ss_cnn_layer}

\end{figure*}

\begin{theorem}
Given a circular input patch $I(a)$ within a larger image, which is defined within the $x,y$ range of $0\leq \sqrt{x^2+y^2} \leq a$. Let $I^s(a)$ denote the same patch when the image was scaled around the centre of the patch by a factor of $s$. We then have
\begin{equation}
   \left[ I^s(a) \star S^{kj}(a) \right] = s^{m-2}e^{-i(k\log s)}\left[ I(as) \star S^{kj}(as) \right]\footnote{This is an exact equality in the continuous domain, but however our signals are discrete, and the extent of deviation will depend on the filter configuration. The deviation is expected to be higher for filters of higher order. To alleviate this error to a certain extent, methods such as input upsampling can be used to improve convolution accuracy. },
\end{equation}
where $\star$ is the cross-correlation operator (in the continuous domain), used in the same context as in \cite{hnet}. 
\end{theorem}
The proof of theorem 1 is shown in the appendix.

An immediate consequence of the above theorem is that for $a=\infty$ the theorem assumes a simpler form, $\left[ I^s \star S^{kj} \right] = s^{m-2}e^{i(k\log s)}\left[ I \star S^{kj} \right]$.

\paragraph{Scale steerability}
A useful consequence of steerability is that any filter expressed as a linear combination (with complex coefficients) of the steerable basis is also steerable. Consider a filter $W(a)$ of radius $a$ constructed in similar fashion using the the proposed scale-steerable basis $S^{kj}$, s.t. $W = \sum_{k,j}{c_{kj}}S^{kj}$, where $c_{kj}\in \mathbb{C}$. The same filter can be steered in its scale by a scale factor of $s$, giving
\begin{equation}\label{eq:scale_steering}
    W^s(as) = s^{m-2}\sum_{k} e^{-ik\log s} \left (\sum_{j}c_{kj}S^{kj}(as) \right). 
\end{equation}
 However, we want the filters to be real valued, and hence we only take the real part of $W^s_{Re}(as)=\Re(W^s(as))$. Note that equality in equation (2) is for both the real and the imaginary parts on both sides of the equation, and thus working with the real part of the filters does not change steerability. The result in Theorem 1 includes an additional change of radius from $a$ to $as$. This indicates that the pixel values of $W^s$ are sampled across a circular region of radius $as$, which depends on the scale factor $s$. Finally, as noted in \cite{steerable_original,hnet}, steerability and sampling are interchangeable, therefore the sampled version of the scaled basis filters are same as the scaled version of the sampled filter. 

\subsection{Scale-Invariant CNNs with Scale Steered Weights}

Here we describe the Scale-Steered CNN (SS-CNN), which employs a scale steeerable filter basis in the computation of its filters. Figure \ref{fig:ss_cnn_layer} shows the proposed scale-invariant layer. Each filter within the scale-invariant layers is computed as a linear combination of the assigned scale steerable basis $S^{kj}$. The network directly only learns the complex co-efficients $c_{kj}$. At each scale-invariant layer, the scaled and resized versions of the filters are directly computed from the complex coefficients using equation \ref{eq:scale_steering}. Only the maximum responses across all scales are channeled to the next layer, by max-pooling the responses across scales.  


\section{Experiments}
First, to validate the proposed approach, datasets such as MNIST-Scale and FMNIST-Scale were chosen which contain global scale variations. In addition, a dataset containing local scale variations was also synthesized from MNIST. Subsequently, the filters and the activation maps within the SS-CNN are visualized. All experiments were run on a NVIDIA Titan GPU processor. The code has been released at \url{https://github.com/rghosh92/SS-CNN}.

\subsection{Classification with SS-CNN}
\subsubsection{MNIST and FMNIST}
The data partitioning protocol for MNIST-Scale is a 10k, 2k, and 50k split of the scaled version of original MNIST, into training, validation and testing data respectively.\footnote{A small training data size is chosen so as to better evaluate the generalization abilities of the trained classifiers.} We use the same split ratio for creating FMNIST-Scale, with the same range of spatial scaling $(0.3,1)$. No additional data augmentation was performed for all the networks.
\paragraph{Global scale variations: MNIST and FMNIST} The results on MNIST-Scale and FMNIST-Scale are shown in Table \ref{tab:global_scale}\footnote{${\star}$ = Our implementation}. The proposed method is compared with three other CNN variants: Locally scale invariant CNN \cite{lsi_cnn}, scale equivariant vector fields \cite{scale_vector_fields} and spatial transformer networks \footnote{For the spatial transformer network, we use network configurations which perform the best on the validation data.} \cite{st_network}. For a fair comparison, all networks used have a total of 3 convolutional layers and 2 fully connected layers. The number of trainable parameters for all four networks were kept approximately the same. Mean and standard deviations of accuracies are reported after 6 splits. \footnote{Note that although the input size for both MNIST and FMNIST are similar, they contain very different kind of data. MNIST is mainly white strokes on a black background, whereas FMNIST includes both shape and texture information in grayscale.}
\begin{table}[H]
\caption{Error rates on MNIST-Scale and FMNIST-Scale}
\label{tab:global_scale}
\scalebox{0.9}{
\begin{tabular}{|c|c|c|}
\hline
                                                        & MNIST-Scale & FMNIST-Scale \\ \hline
Spatial Transformer \cite{st_network}                                   & 1.97$\pm$0.09$^{\star}$  & \textbf{13.11}$\pm$\textbf{0.25}$^{\star}$   \\ 
\begin{tabular}[c]{@{}c@{}}SS-CNN (Ours)\end{tabular} & \textbf{1.91}$\pm$\textbf{0.04}  & 14.24$\pm$0.31   \\ 
LocScaleEq-CNN \cite{scale_vector_fields}                                            & 2.44$\pm$0.07  & 15.72$\pm$0.32$^{\star}$   \\ 
LocScaleInv-CNN \cite{lsi_cnn}                                    & 2.75$\pm$0.09  & 15.91$\pm$0.41$^{\star}$   \\ \hline
\end{tabular}}
\end{table}
\vspace{-30pt}

\paragraph{Generalization to Distortions} Here we test and compare method performance on MNIST with added elastic distortions. The networks are all trained on the undistorted MNIST-Scale, but tested on MNIST-Scale with added elastic deformations.  Results are shown in Table \ref{tab:mnist_scale_elastic}. We only record the performance for a single network (best performing) for each method.  
\vspace{-20pt}

\begin{table}[H]
\caption{Results: Test-time Elastic Distortions on MNIST-Scale}
\label{tab:mnist_scale_elastic}
\scalebox{0.9}{
\begin{tabular}{|c|c|c|c|c|c|}\hline
                                                        & $\alpha$=0 & $\alpha$=10 & $\alpha$=20 & $\alpha$=30 & $\alpha$=40 \\ \hline
ScaleInv Net                                            & 3.2    & 5.92    & 9.6     & 16.2     & 27       \\ 
Spatial Transformer                                     & 1.87   & 3.4   & 5.12    & 9.2     & 16.2     \\ 
\begin{tabular}[c]{@{}c@{}}SS-CNN (Ours)\end{tabular} & 1.87   & 3.7     & 5.6     & 9.82    & 16.83    \\ 
\hline
\end{tabular}}
\end{table}
\vspace{-30pt}

\paragraph{Synthesized data: Local scale variations} 
We synthesize a variation using MNIST, namely MNIST-scale-local-2, with scale variations that are more local than MNIST-Scale. Pairs of MNIST examples were each scaled with a random scale factor between $(0.7,1)$, and arranged side by side in an image of size $28\times40$, a small proportion of which contains overlapping examples. We only choose 10 possible combinations of digits, $(0,1),(1,2),(2,3),(3,4),..,(9,0)$, resulting in a total of 10 categories for the network.  Mean and standard deviations of accuracies are reported after 6 splits. Results are reported in Table \ref{tab:mnist_scale_local}. The results demonstrate the superior performance of local scale-invariance based methods over global transformation estimation architectures such as spatial transformers, in a scenario where the data contains local scale variations. 

\begin{table}[H]
\caption{Results on MNIST-scale-local-2: Varying Training Data Size}
\label{tab:mnist_scale_local}
\scalebox{0.9}{
\begin{tabular}{|l|l|l|l|}
\hline
                    & 1\% data           & 10\% data         & 100\% data         \\ \hline
Spacial Transformer & 4.76$\pm$0.38          & 0.73$\pm$0.05         & 0.23$\pm$0.02          \\ 
SS-CNN(ours)        & \textbf{4.27}$\pm$\textbf{1.14} & \textbf{0.4}$\pm$\textbf{0.02} & \textbf{0.09}$\pm$\textbf{0.01} \\ \hline
\end{tabular}}
\end{table}
 \vspace{-30pt}

\subsection{Visualization Experiments}

\begin{figure}[h]
    \centering
    \includegraphics[width=0.35\textwidth]{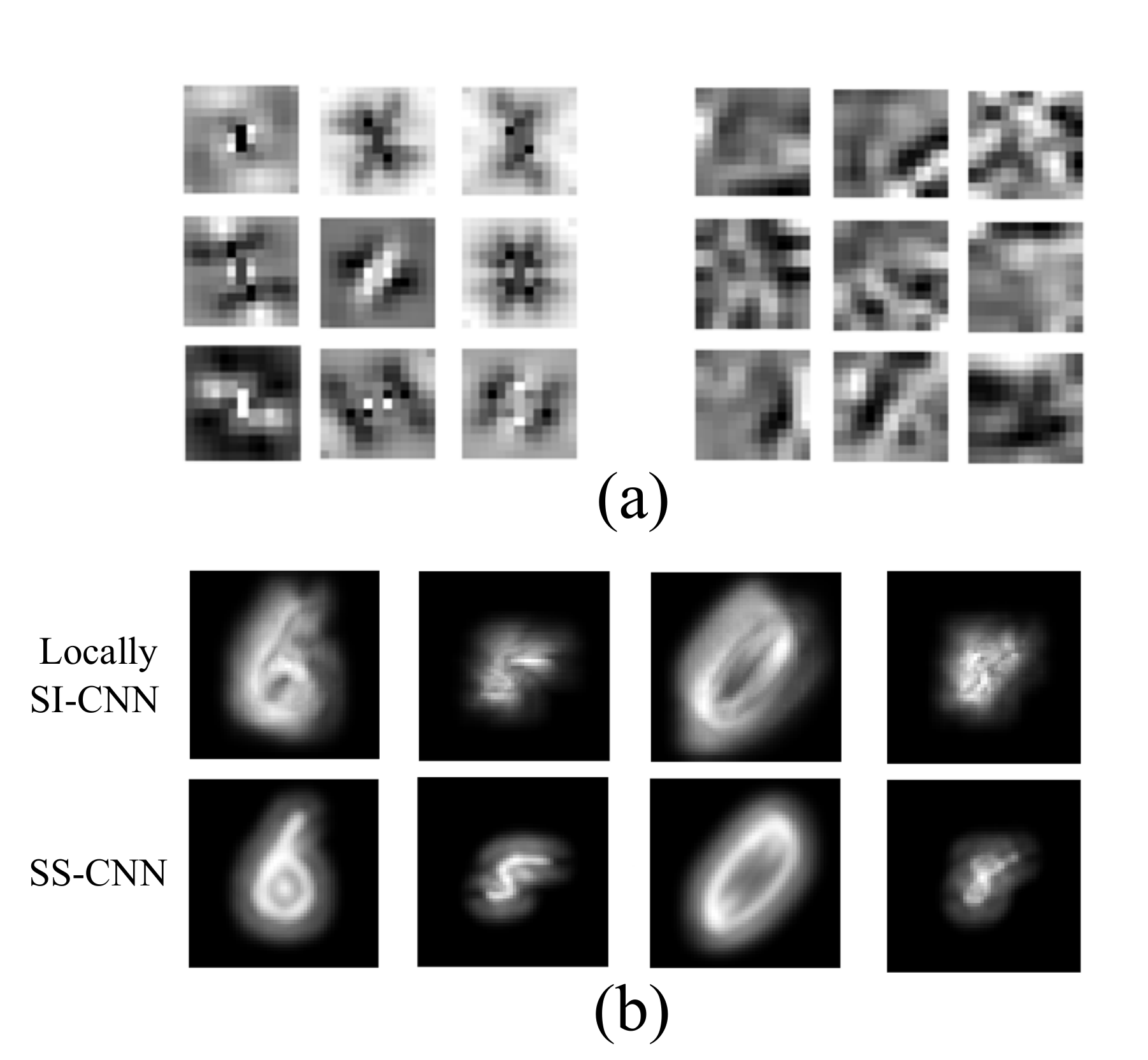}
    \caption{Visualized Filters and Average Feature Map Activation: (a) shows first layer filters generated from MNIST-Scale training for the proposed SS-CNN (left) and the conventional locally scale-invariant CNN (right), and (b) shows the average feature map activation of the first layer output of the LocScaleInv-CNN (top) and the SS-CNN (bottom). }
    \label{fig:filters_n_maps}
\end{figure}

In this section we visualize the network filters and feature map activations for two scale-invariant networks: our proposed SS-CNN and the LocScaleInv-CNN. Both networks were trained on MNIST-Scale. Figure \ref{fig:filters_n_maps} (a) shows a visual comparison of the layer 1 filters for these networks. Notice that the scale-steered filters show considerably higher structure, centrality, and interesting filter form: some of them resembling oriented bars. Figure \ref{fig:filters_n_maps} (b) compares the average feature map activation of Layer 1, in response to different inputs. Notice that spatial structure is far better preserved in the SS-CNN responses (bottom row), with the digit outlines clearly distinguishable. This is partly due to the ingrained centrality of the scale-steered basis (the $1/r$ term), which generates a response which is more structure preserving. 

\section{Discussions}

Based on the proposed SS-CNN framework in this work, we underline some of the important issues and considerations moving forward. Also, we provide detailed explanations for some of the design choices used in this work. 

\begin{itemize}
    \item \textbf{Input Resizing vs Filter Scaling:}
    For locally scale invariant CNNs, usually the input is reshaped to a range of sizes, both smaller and greater than the original size \cite{lsi_cnn,scale_vector_fields}. Feature maps are obtained by convolving each resized input with an unchanged filter. Lastly all the feature maps are reshaped back to a common size, beyond which only the maximum response across scales are channeled. This approach uses two rounds of reshaping, and thus is clearly prone to interpolation artifacts, especially if the filters are not smooth enough. The method proposed in this work only steers the filters in their scale and size, without having to rely on any interpolation operations. Note that change of filter size just requires computing the filter values at the new locations using equation \ref{eq:scale_steering} and \ref{eq:steerable_form}.
    \item \textbf{Filter Centrality:} 
    If the filters are not central, i.e. centered near to their centre of mass\footnote{Centre of mass, in this case holds the same definition as in physics. The "mass" element can be considered as the absolute value of the filter at a certain location.}, then they pose the risk of entangling scale and translation information. This happens, when the filter response to the input, at a certain scale and location is the same as the response of the same filter at a different scale and a different location. This can be quite common for filters which have most of their "mass" away from their center. Such entanglement can often lead to feature maps with distorted and over-smoothed spatial structure, as observed in Figure \ref{fig:filters_n_maps} (b) (top). This issue can be tackled to a certain extent by using filters which show centrality (Figure \ref{fig:filters_n_maps} (a)). As seen in equation \ref{eq:steerable_form}, one can control the centrality of the steerable basis filters, with the radial term $(1/r^m)$, and by ensuring radial-symmetric filters with $(K(\phi,\phi_j)+K(\phi,\phi_j+\pi))$ as the angular term. Figure \ref{fig:steerable_basis} shows the central nature of the steerable basis.  Filter centrality is preserved for the subsequently generated filters, as seen in Figure \ref{fig:filters_n_maps} (a) (left), which shows the generated filters after training. 
    \item \textbf{Transformation Sensitivity:} As iterated in section \ref{sec:intro}, an important yet partly overlooked aspect of using a steerable basis from the family of circular harmonics (or log-radial harmonics), is the ability to control the transformation sensitivity of the filters. For instance, circular harmonics beyond a certain order have a much smaller sensitivity to changes in input rotation. This is simply because each circular harmonic filter is invariant to discrete rotations of $2\pi/k$, $k$ being the filter order. Similarly, it is easily seen that each log-radial harmonic filter is invariant to filters being scaled by a scale factor of $e^{\pm 2\pi/k}$. Therefore, higher order filters show considerably less transformation sensitivity. It is perhaps noteworthy that the 2D Fourier transform (or the 2D DCT) basis functions can also be used as a steerable basis (e.g. \cite{spectral_cnn}). In that case, higher frequency (analogous to filter order) filters are less sensitive to input translations, compared to low frequency filters. Therefore in a certain sense, the circular harmonic and log-radial harmonic filter bases are a natural extension of the Fourier basis (translations), to other transformations (rotation and scale).

\end{itemize}

\section{Conclusions and Future Work}
A scale-steerable filter basis is proposed, which along with the popular rotation-steerable circular harmonics, can help augment CNNs with a much higher degree of transformational weight-sharing. Experiments on multiple datasets showcasing global and local scale variations demonstrated the performance benefits from using scale-steered filters in a scale-invariant framework. Scale-steered filters are found to showcase heightened centrality and structure. A natural trajectory for this approach will be to inculcate the scale-steering paradigm onto equivariant architectures such as GCNNs.

\section*{Acknowledgments}
This research was supported by DSO National Laboratories, Singapore (grant no. R-719-000-029-592). We thank Dr. Loo Nin Teow and Dr. How Khee Yin for helpful discussions. We also thank Dr. Diego Marcos for sharing the code for scale-vector fields \cite{scale_vector_fields}, and clarifying a number of other queries related to the task.


\bibliography{all_refs}
\bibliographystyle{ieeetr}

\appendix
\section{Proof of Theorem 1}
\begin{proof}
First, note that in the log-polar domain, $dxdy=rdrd\phi=r^2d(\log r)d\phi$. Using this fact, the cross-correlation $\left[ I^s(a) \star S^{kj}(a) \right]$ can be expressed in log-polar co-ordinates as the integration
\begin{equation}
    \int_{-\infty}^{\log a} \int_{0}^{2\pi} I(z + \log s,\phi)S^{kj}(z,\phi)e^{2z}dzd\phi,
\end{equation}

where $z=\log r$.  A change of integrands from $dz$ to $dz'$, where $z'= z + \log s$, yields 
\begin{equation}
    \int_{-\infty}^{\log a + \log s} \int_{0}^{2\pi} I(z',\phi)S^{kj}(z'-\log s ,\phi)\frac{e^{z'}}{s^2}dz'd\phi.
\end{equation}
From the definition of the steerable filter basis $S^{kj}$, we have that $S^{kj}(z'-\log s ,\phi)=s^{m} \times S^{kj}(z',\phi)e^{-ik\log s}$. Thus the integration can be further simplified as, 
\begin{align}
    &s^{m-2}e^{-ik\log s} \int_{-\infty}^{\log a + \log s} \int_{0}^{2\pi} I(z',\phi)S^{kj}(z' ,\phi)e^{2z'}dz'd\phi  \\
    &= s^{m-2}e^{-i(k\log s)}\left[ I(as) \star S^{kj}(as) \right].
\end{align}
This completes the proof.
\end{proof} 

\section{Steerable Basis Parameters} 

The definition of each log-radial harmonic filter includes a total of four parameters: phase ($\beta$), filter order ($k$), filter orientation $\phi_j$ and orientation spread ($\sigma_{\phi}$). For all networks that have been trained in this work using scale-steered filters, we keep $\beta=0$, $k=(0.5,1,2)$, $\phi_j=j(\pi/8),j \in [1,8]$ and $\sigma_{\phi}=\pi/16$. Note that this configuration of the steerable basis space leads to a total of 24 log-radial harmonics as the steerable basis. Thus, each scale-steerable filter has $24*2=48$ trainable parameters (Due to both real and imaginary components on each coefficient). One additional aspect of note is the $\log r$ term in the complex exponential $e^{ik(\log r)}$ in the filter definition. Since the filter is undefined for $r=0$, we enforce $S^{k,j}(r,\phi)=1$ for $r=0$.      

\section{Network Configuration Used}

In each layer of the SS-CNN the filter scale factors are within the range $(1,2.4)$, with the size of the filters increasing from $(7,7)$ to $(17,17)$ (only odd size filters are chosen because of well defined centre pixel). For such large filter sizes, an additional upsampling of factor 2 was applied on the data. Note that upsampling ensures more precise convolutions, especially with scale-steered filters of higher orders. Note that although upsampling adds slight improvements to the SS-CNN ($\approx 0.2\%$ in MNIST-Scale), we found that it does not improve the performance of the other networks compared in this paper. For all experiments, the number of feature maps of within each layer were (30,60,90), for all networks. A total of 3 max-pooling layers were used after the first ($2\times2$), second ($2\times2$) and the third convolution layer ($8\times8$ for the SS-CNN, $4\times4$ for other networks). For the FMNIST-Scale and MNIST-Scale-local it was ensured that all networks had approximately the same number of trainable parameters. All networks were trained for a maximum of 300 epochs, after which the best performing model on the validation data was used for testing. No data augmentation was used in any experiment.






\end{document}